# Solving the Satisfiability Problem Through Boolean Networks

Andrea Roli and Michela Milano

DEIS, Alma Mater Studiorum Università di Bologna

andrea.roli, michela.milano @unibo.it

**Abstract.** In this paper we present a new approach to solve the satisfiability problem (SAT), based on boolean networks (BN). We define a mapping between a SAT instance and a BN, and we solve SAT problem by simulating the BN dynamics. We prove that BN fixed points correspond to the SAT solutions. The mapping presented allows to develop a new class of algorithms to solve SAT. Moreover, this new approach suggests new ways to combine symbolic and connectionist computation and provides a general framework for local search algorithms.

## 1 Introduction

The satisfiability problem (SAT) [Garey and Johnson, 1979] has an important role in computer science and it has been widely investigated. The SAT problem is a NP-complete problem concerning the satisfiability of boolean formulas, i.e., find an assignment of boolean values to variables such that the formula is satisfied. SAT is very important in several Artificial Intelligence areas, like propositional calculus, constraints satisfaction and planning. For its theoretical and practical relevance, many specialized (complete and incomplete) algorithms have been developed.

We present a novel approach to solve SAT, based on Boolean Networks (BN). Up to this time, boolean networks have been used for modeling complex adaptive systems [Cowan *et al.,* 1994], and in the field of machine learning (see for instance [Dorigo, 1994]). In this approach, we map a SAT instance in a BN, and we simulate its dynamics; the stationary states of the net correspond to the solutions of SAT. Due to the BN structure and dynamics used, the resulting algorithms are incomplete.

We have developed and tested three algorithms, each of them is derived from a variant of boolean networks: synchronous, probabilistic and asynchronous boolean nets. The first algorithm has led to disappointing results, while the second and the third had performed better.

The new approach represents a bind between symbolic and connectionist computation, and it allows to develop new algorithms to solve SAT.

This work represents a first investigation on the subject and it mainly refers to the founding principles. The algorithms presented are based on elementary dynamics of boolean networks, without using any heuristic function to guide the search, nor optimization criteria.

## 2 Preliminaries

In this section we recall the boolean network model (for more details see for instance [Kauffman, 1993]). Then, we briefly define the satisfiability problem and recall some algorithms.

### 2.1 Boolean Networks

Boolean networks (BN) have been introduced by Kauffman [Kauffman, 1993] as a model of genetic networks (models of genes activity and interactions) and as a framework to study complex adaptive systems (see [Kauffman, 1993; Cowan *et al.*, 1994]).

A BN is a directed graph of *n* nodes; each node *i* of the graph is associated with a boolean variable ($v_i$) and a boolean function ($F_i$). The inputs of the boolean function $F_i$ are boolean variables associated with the neighboring nodes (i.e., nodes whose outgoing arcs lead to the node *i*). See, for instance, fig.1 (left part).

The network is initialized with a random or deterministically selected initial state; the dynamics of the net is given by the state transition of the nodes (see fig.1 right part), depending on the results of the corresponding boolean functions. In the main definition of boolean networks the dynamics is synchronous, i.e., nodes are updated in parallel. We will consider, also, asynchronous dynamics, i.e., nodes are sequentially updated, and probabilistic dynamics, i.e., each node has a set of boolean functions and it is updated choosing one of them.

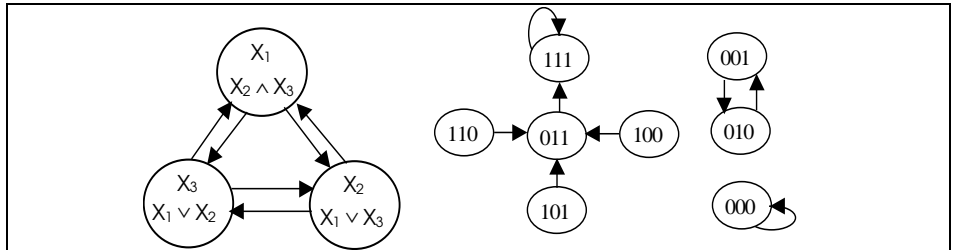

**Fig.1** A synchronous boolean network and its trajectories

A BN is a discrete-time dynamic system with boolean variables; we can analyze it observing the *trajectory* in the state space, the length of the *transient phase*, the type of the *attractors* and the *stability*. The state of the system is given by the array $\mathbf{x}=(x_1,...,x_n)$, $x_i \in \{0,1\}$, $1 \leq i \leq n$.

Since the number of states is finite ($2^n$) and the transition rules are deterministic, eventually the system will reach a state already found, that is, after a transient, it will find a *cyclic attractor* and it will repeat the same sequence of states (*cycle of states*). The number of states constituting the cycle represents the *length* (*period*) of the attractor; the minimum period is 1 (i.e., the attractor is a *fixed point*) and the maximum period is the number of states of the system ($2^n$).

The dynamics of the network is given by the transition rules: $x_i(t+1)=F_i(x_{i_1}(t),...,x_{i_{k(i)}}(t))$, where $x_{i_j}$ (j = 1,2,...,k(i)) are the inputs of the boolean function $F_i$, for i = 1,2,...,n.

We define *basin of attraction* of an attractor the collection of states such that, if selected as initial states, will converge to the attractor. See fig.1 for an example of a boolean network dynamics.

### 2.2 The Satisfiability Problem

The satisfiability problem (SAT) is a well-known NP-complete problem (see [Garey and Johnson, 1979]) and it has an important role in computer science; in fact, it is possible to transform every NP problem in SAT in polynomial time. Furthermore, many applications (e.g., planning, VLSI testing, Boolean CSP) can be expressed in terms of SAT.

We will refer to the following definition of SAT: given a boolean expression in conjunctive normal form (CNF), i.e., a conjunction of clauses, each of them constituted by disjunction of variables, negated or not, find an assignment of boolean variables which satisfies the expression. For example, consider the expression $\Phi = (x_1 \vee x_2 \vee \sim x_3) \wedge (\sim x_2 \vee x_3) \wedge (\sim x_1 \vee \sim x_2 \vee \sim x_3) \wedge (\sim x_1 \vee x_2)$; $\Phi$ is constituted by four clauses, each of them contains the disjunction of some *literals* (i.e., positive or negative variables). Given a truth assignment $T$ (that is, we assign the value 'True' (1) or 'False' (0) to each variable) we say that $\Phi$ is satisfied by $T$ if and only if every clause contains at least one literal with the 'True' value. In this case, $\Phi$ is satisfied by the assignments $\{x_1=0, x_2=0, x_3=0\}$ and $\{x_1 = 0, x_2 = 1, x_3 = 1\}$.

Two kinds of algorithms for solving SAT have been proposed in the literature: complete and incomplete algorithms. Complete algorithms always find a solution, if it exists, in finite time; incomplete algorithms could not find a solution, even if it exists. Among complete algorithms, procedures derived from the Davis-Putnam (DP) algorithm [Davis and Putnam, 1960] are the most efficient. Despite the guarantee of finding a solution, complete algorithms are seldom used for real applications, because they are computationally inefficient. In recent years, some incomplete algorithms have been developed (*model finders*); among the most efficient incomplete algorithms we mention GSAT [Selman *et al.*, 1992], WalkSAT [McAllester *et al.*, 1997] and MSLSAT [Liang and Li, 1998]. Incomplete algorithms are widely used, because they are much more efficient then complete ones and, on average, they can solve most of the satisfiable instances.

## 3 Solving SAT Problems with BNs

In this section, we present a new approach to solve the satisfiability problem, which consists in the transformation of a SAT problem in a BN. For this purpose we define a *mapping* that generates, given a SAT instance, a BN whose dynamics is characterized by the presence of fixed points corresponding to the solutions of the SAT instance. Then we simulate the dynamics of the network until a steady state is reached. The steady state represents the solution of the problem. The algorithms derived are incomplete. We are currently investigating if, by changing the mapping and dynamics, we can obtain complete procedures.

### 3.1 The Mapping

The core of the application is the particular transformation that allows to switch from the symbolic space of the propositional satisfiability to the BN sub-symbolic space, preserving the correctness of the results.

The fundamental requirement of a mapping is to give a correspondence between the solutions of the SAT instance and the BN fixed points. More precisely, if we as-sume a one-to-one correspondence between the boolean variables of the propositional formula and the nodes of the network, the mapping should be such that a satisfying assignment for the formula corresponds to a fixed point of the network (*completeness* property); moreover, every fixed point of the boolean network corresponds to a solution of the SAT instance (*soundness* property).

We designed a simple mapping (called $\mu 1$) which has the desired requirements; the BN has $n$ nodes if $n$ are SAT variables. For each variable, the corresponding boolean function is computed as follows:

**Input**: $\Phi = c_1 \wedge c_2 \wedge ... \wedge c_m$ (a boolean formula in CNF);
**Output**: Boolean functions $F_i$ ;
**For each variable** $x_i$ **define** $O_i = \{c_j, j = 1,..,m \mid x_i \in c_j\}$, $A_i = \{c_j, j = 1,..,m \mid \sim x_i \in c_j\}$
**For** i:=1 **to** n **define** $F_i = (x_i \wedge And[A_i]) \vee \sim And[O_i]$

where the function «*And*» is the logical operator '$\wedge$' applied to the elements of $A_i$, being '1' the result of its application to an empty set.

For example, consider the following SAT instance: $\Phi_1 = (x_1 \vee \sim x_2) \wedge (\sim x_1 \vee x_2) \wedge (x_2 \vee x_3)$ = $c_1 \wedge c_2 \wedge c_3$; we use a boolean network of three nodes with $(x_1, F_1)$, $(x_2, F_2)$, $(x_3, F_3)$. The sets $O_i$ and $A_i$ are: $O_1 = \{c_1\}$, $A_1 = \{c_2\}$; $O_2 = \{c_2, c_3\}$, $A_2 = \{c_1\}$; $O_3 = \{c_3\}$, $A_3 = \varnothing$. The boolean functions are: $F_1 = (c_2 \wedge x_1) \vee \sim c_1$, $F_2 = (c_1 \wedge x_2) \vee \sim c_2 \vee \sim c_3$, $F_3 = x_3 \vee \sim c_3$. Note that $\Phi_1$ is satisfied iff $(x_1, x_2, x_3) \in \{(0,0,1), (1,1,0), (1,1,1)\}$ representing the only fixed points of the network.

It is possible to prove that the mapping $\mu 1$ is sound and complete.

**Proposition 1**: given an instance $\Phi$ of SAT (with $n$ variables), the boolean network $\mathfrak{R}$ of $n$ nodes induced by the mapping $\mu 1$ is such that its fixed points are in one-to-one correspondence with the solutions of $\Phi$ (see Appendix A for the proof).

Note that the mapping $\mu 1$ can be efficiently implemented. If the SAT has $n$ variables, m clauses and $l_{max}$ is the maximum number of literals per clause, the definition of $A_i$ and $O_i$ can be executed in time $O(m\, l_{max})$.

### 3.2 The Dynamics of the Network

The second phase of the algorithm is the simulation of the dynamics of the network. The boolean net associated (by means of $\mu 1$) with the problem is now the dynamic system that performs the computation. The goal of the simulation is to find a fixed point. Note that the mapping obeys only the condition about fixed points, nothing has been imposed about cycles. If during the simulation of a synchronous and deterministic dynamics the network reaches a cycle, it has to be reinitialized and restarted with a new trajectory in the state space. In order to avoid deterministic cycles, we have investigated also probabilistic and asynchronous BNs, for which we proved that (deterministic) cycles do not exist.

We have tried three different kind of dynamics, i.e., synchronous, probabilistic and asynchronous, while maintaining the same mapping ($\mu 1$). In the next subsections we will describe each of them.

#### 3.2.1 Synchronous Boolean Networks

Given a SAT instance we apply $\mu 1$ to obtain a boolean network; in a Synchronous Boolean Network (hereinafter referred to as SBN), variables are updated in parallel and transitions are deterministic.

Example: given the boolean formula $\Phi_2 = (x_1 \vee x_2) \wedge (x_1 \vee \sim x_3) \wedge (\sim x_1 \vee \sim x_3) \wedge (x_2 \vee x_3) \wedge (x_1 \vee x_3) = c_1 \wedge c_2 \wedge c_3 \wedge c_4 \wedge c_5$, $\mu 1$ generates a boolean network with three nodes, de-fined by these functions: $F_1 = (c_3 \wedge x_1) \vee \sim c_1 \vee \sim c_2 \vee \sim c_5$, $F_2 = x_2 \vee \sim c_1 \vee \sim c_4$, $F_3 = (c_2 \wedge c_3 \wedge x_3) \vee \sim c_4 \vee \sim c_5$. The solution of $\Phi_2$ is $(x_1, x_2, x_3) \in \{(1,1,0)\}$ and corresponds to the transition graph fix point depicted in fig.2.

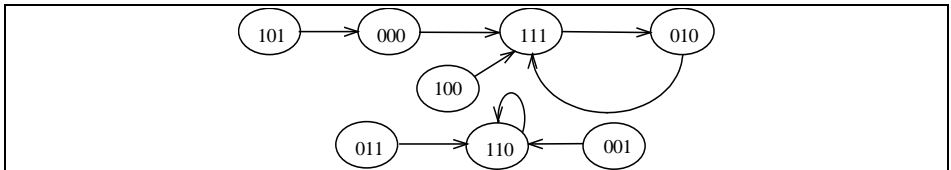

**Fig. 2.** Transition graph of the BN associated with $\Phi_2$: the fixed point has a basin of attraction of 3 states. There is also a cycle which has period 2 and a basin of attraction of 5 states.

The algorithm for SBN, depicted in figure 3, simulates the network dynamics until a fixed point is found or the maximum number of iterations is reached. If the network is trapped in a cycle, the procedure restarts the network from a new random initial state. SAT-SBN1 detains disappointing performances: it gave not better performance than a *generate and test* procedure. Experimental results suggest that the lower is the number of satisfiable assignments, the lower the size of the fixed points basin of attraction.

Note that the inefficiency of the algorithm comes from the combination of the specific mapping with synchronous dynamics and it is not an intrinsic characteristic of the approach.

```
procedure SAT-SBN1;
begin
  iterations := 1;
  while (iterations ≤ MAX_ITER) do
  begin
    attractor := False; trajectory := [];
    select_random_initial_state;
    while (attractor==False && iterations≤ MAX_ITER) do
    begin
      compute_new_state;
      Append(trajectory, state);
      if end_of_transient(trajectory) then attractor := True
        else iterations++;
    end;
    if fixed_point(trajectory) then return solution;
    else iterations++;
  end;
end;
```

**Fig. 3.** SAT-SBN1 algorithm

### 3.2.2 Probabilistic BNs

One way to avoid cycles is to introduce a non-deterministic system transition function where a system state has more than one successor state, each one with a given probability.

The probabilistic version of a boolean network (PBN) is obtained by associating more than one boolean function with each node and specifying a *function probability*, which gives the probability of selecting a particular boolean function. Each state transition is obtained by selecting one boolean function per node and synchronously updating all nodes involved in the selected function. Since for each state several transitions are possible, the transition graph has nodes with more than one outcoming arc and each arc has a transition probability.

A fixed point for a PBN transition graph is a node with a self-arc whose transition probability is equal to 1. PBN can present *(probabilistic) cycles*, which are cycles composed by arcs with a probability less than 1. We call *deterministic cycles* those whose arcs have transition probability equal to 1.

In this paper, we consider PBNs obtained by generating a SBN with $\mu 1$ and adding to each node an identity boolean function ($F_i = x_i(t+1) = x_i(t)$); the transition probability of the first boolean function is p, and for the identity function is 1-p. Thus, each node changes its value according to the original boolean function with probability p, and maintains the same value with probability 1-p.

The algorithm that simulates the dynamics of the PBN is reported in figure 4. The algorithm must recognize true fixed points, distinguishing them from repetition of the same state, even if it is not a fixed point. This is done by verifying if the current assignment (i.e., the current state) satisfies the original formula (see the statement "`if (satisfied_formula(state)) then return solution`" in fig.4), or by executing a SBN1-like step. Even if this operation has a computational high cost if frequently executed, SAT-PBN1 strongly outperformed SAT-SBN1. We experimentally found that an *optimal probability p* exists for which the algorithm gets the best performance; for 3-SAT *p* is near 0.2.

```
procedure SAT-PBN1;
begin
  iterations := 1;
  select_random_initial_state;
  while (iterations ≤ MAX_ITER) do
  begin
    old_state := state;
    compute_new_state_with_transition_prob_p;
    if (old_state == state) then
      if (satisfied_formula(state) ) then return solution;
    iterations++;
  end;
end;
```

**Fig. 4** SAT-PBN algorithm

PBN1 has the following property:
**Proposition 2:** If the network is generated by means of $\mu 1$ from a satisfiable boolean formula, for every initial state the probability that the network reaches a fixed point tends to 1; that is:

$$\lim_{time \to \infty} Prob\{\text{"the net reaches a fixed point"}\} = 1$$

(see Appendix B for a proof). As a consequence of the previous proposition PBNs of this application are "deterministic cycle – free".

Proposition 2 allows to formally define the convergence of the algorithm SAT-PBN1 in terms of *Probabilistic Asymptotic Completeness* [Hoos, 1999] asserting that the algorithm SAT-PBN1 is Probabilistically Approximately Complete (PAC).

### 3.2.3 Asynchronous BNs

Asynchronous boolean networks (ABN) are characterized by the sequential update of the nodes. There are several ways to update the nodes, either fixed or random sequences or sequences obtained by other kinds of probabilistic distributions. In this work, we use random update sequences: at each iteration only one randomly selected node is updated. The algorithm for ABN (SAT-ABN1) is depicted in fig.5. Since the computational cost of the "true fixed point" test is high, we structured the update sequences in this way: the dynamics of the network is divided in *macro-transitions*, which are random sequential update of all the N nodes (each single update is called *micro-transitions*). Since, if a state is a fixed point, every micro-transition is such that the variable maintains the old value, after a macro-transition the fixed point is correctly recognized. Vice versa, a simple repetition of a state, which is not a fixed point, is possible only during the macro-transition (that is: between micro-transitions) and, after the macro-transition, the new state is surely different from the old one. The asynchronous dynamics allows a kind of *communication* between the nodes: since they are updated one at a time, two or more variables do not change their value to satisfying the same clause. The use of macro-transitions gains the performance of the SAT-ABN1 algorithm.

```
procedure SAT-ABN1;
begin
  iterations := 1;
  select_random_initial_state;
  while(iterations≤ MAX_ITERATIONS) do
  begin
    old_state := state;
    compute_new_state_with_random sequence;
    if (old_state == state) then return solution;
    else iterations++;
  end;
end;
```

**Fig. 5.** SAT-ABN1 algorithm

Proposition 2, presented in the previous subsection, is also valid in the ABNs case. Therefore, we can assert that ABNs of this application are "deterministic cycle-free" and the algorithm SAT-ABN1 is PAC.

### 3.3 Experimental Results

We compared the BN-based algorithms on 3-SAT random generated satisfiable (*forced*) formulas with $n$ variables and $m$ clauses. Since the synchronous version showed non competitive performances, even for 20 variables, we tested only SAT-PBN1 (with $p$=0.2) and SAT-ABN1. The run time was limited and a negative result was reported if a solution was not found. ABN and PBN were restarted after a number of transitions proportional to $n^2$. In Tables 1,2 are shown samples of the experimental results. The algorithms have been implemented in C and run on a PentiumII 233 Mhz.

SAT-PBN1 is competitive with SAT-ABN1 for $n>500$ and $m<3n$. When $m/n<3$, 3-SAT instances have many solutions and the parallel search is most efficient. For $m/n>3$ more conflicting constraints have to be satisfied and the sequential search works better.

|   |   | SAT-ABN1 | | | SAT-PBN1 | | | GSAT | | |
|---|---|---|---|---|---|---|---|---|---|---|
| n | m | time (msec) | iter. | solved | time (msec) | iter. | solved | time (msec) | iter. | solved |
| 50 | 100 | 10 | 3 | 100% | <1 | 22 | 100% | <1 | 11 | 100% |
| 50 | 150 | 10 | 9 | 100% | 10 | 156 | 100% | <1 | 26 | 100% |
| 50 | 215 | 310 | 894 | 99% | 3155 | 34073 | 88% | 10 | 105 | 100% |
| 80 | 160 | 20 | 4 | 100% | <1 | 29 | 100% | <1 | 22 | 100% |
| 80 | 240 | 20 | 14 | 100% | 30 | 257 | 100% | 10 | 72 | 100% |
| 80 | 344 | 9113 | 15958 | 55.5% | 13550 | 97891 | 12.5% | 30 | 428 | 100% |
| 100 | 200 | <1 | 4 | 100% | 10 | 34 | 100% | <1 | 33 | 100% |
| 100 | 300 | 20 | 19 | 100% | 90 | 598 | 100% | 10 | 97 | 100% |
| 100 | 430 | 21681 | 28526 | 16.5% | - | - | 0% | 60 | 590 | 100% |
| 200 | 400 | 20 | 5 | 100% | 30 | 50 | 100% | 20 | 94 | 100% |
| 200 | 600 | 121 | 64 | 100% | 1011 | 3664 | 100% | 60 | 287 | 100% |
| 200 | 860 | - | - | 0% | - | - | 0% | 471 | 2435 | 100% |

**Table 1**
Median execution time and iterations over 200 satisfiable instances.

We also compared these procedures with GSAT, but we obtained disappointing results: GSAT is faster and more effective than the BN-procedures. This is due to the fact that GSAT is based on heuristic criteria, which guide the search, while the simple BN-procedures perform a "blind" search. Nevertheless, even without heuristic, BN-procedures perform better than GSAT when $m<3n$, as we can see from Table 2. The number of GSAT maxflips was fixed to $5n$ (according to [Selman *et al*., 1992]).

We also tested the procedures on random non-forced instances and we observed the same qualitative behavior (in this case, BN-procedures perform better than GSAT for $m<2.5n$).

|   |   | SAT-ABN1 | | | SAT-PBN1 | | | GSAT | | |
|---|---|---|---|---|---|---|---|---|---|---|
| n | m | time (msec) | iter. | solved | time (msec) | iter. | solved | time (msec) | iter. | solved |
| 300 | 600 | 41 | 6 | 100% | 50 | 57 | 100% | 61 | 174 | 100% |
| 300 | 750 | 90 | 14 | 100% | 120 | 180 | 100% | 100 | 253 | 100% |
| 500 | 1000 | 140 | 7 | 100% | 130 | 70 | 100% | 201 | 363 | 100% |
| 500 | 1250 | 220 | 18 | 100% | 231 | 225 | 100% | 330 | 722 | 100% |
| 700 | 1400 | 250 | 9 | 100% | 190 | 76 | 100% | 391 | 601 | 100% |
| 700 | 1750 | 450 | 23 | 100% | 390 | 282 | 100% | 721 | 1186 | 100% |
| 1000 | 2000 | 481 | 9 | 100% | 341 | 84 | 100% | 851 | 914 | 100% |
| 1000 | 2500 | 871 | 27 | 100% | 670 | 343 | 100% | 1432 | 1688 | 100% |

**Table 2**
Median execution time and iterations over 100 satisfiable instances.

### 3.4 Discussion

The asynchronous version of the algorithm (with µ1) is analogous to a local repair algorithm; in fact, variables that belong to unsatisfied clauses are forced to change their value. The WalkSAT algorithm, with random choice of the variable within an unsatisfied clause, is indeed very similar to ABN-based algorithms. Furthermore, SAT-ABN1 is a kind of WalkSAT with random choice and variable-length tabu list. Main differences are:

   - ABN-based algorithms are intrinsically concurrent and, when sequentialized, they update even those variables which belongs to satisfied clauses; this implies many "void updates", which decrease the performance (notice that SAT-PBN1 can be viewed as a sort of parallelized version of WalkSAT).

   - WalkSAT shows the best performance with heuristic function which guides the search; such heuristic is completely absent in SAT-ABN1. We are currently working on the introduction of heuristic knowledge in BN-algorithms.

Finally, it is worth noticing that the boolean functions defined by µ1 explicitly make the *pure literals* simplification (only in the first processing phase), since they fix the values 1 (0) to those updating variables which compare only non-negated (negated).

The combination of functional computation and dynamics can be generalized using a General Framework (see fig.6). Moving along the two dimensions (which represent the functional/dynamics complexity) it is possible to design new algorithms (like SAT-ABN1) and redesigning old ones. For example, with a particular choice of dynamics and mapping, local search algorithms, like GSAT, WalkSAT and their variants can be reinterpreted in the BN framework. The mapping in this case creates boolean functions imposing a simple flip of the updating variable, $F_i = x_i(t+1) = \sim x_i(t)\ \forall i$. The search mechanism is performed by the asynchronous (sequential) dynamics which selects the updating variable according to the heuristic criteria of the search procedures.

This framework allows compared analysis and generalization of local search and local repair procedures. A first result of this approach is that, as a consequence of Proposition 2, GSAT and WalkSAT with noise can find a solution with probability 1 (with unlimited time).

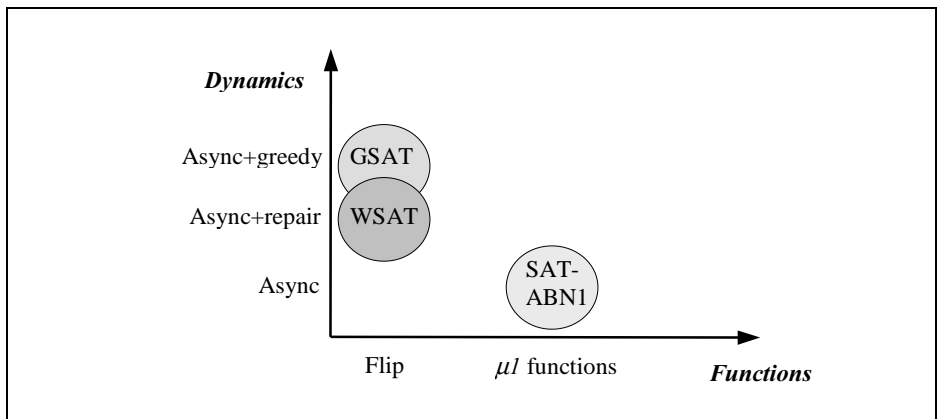

**Fig. 6** The separation between functions and dynamics allows to consider old and new algorithms in a unique framework.

## 4 Conclusion and Future Work

In this paper we have introduced a new approach to the solution of SAT, based on boolean networks. The transformation from a SAT instance to a boolean network is allowed by a mapping, which is sound and complete. The dynamics of the SAT-generated networks corresponds to the computation phase and the stationary state of the system is the solution of the problem.

We designed a mapping ($\mu 1$) and we developed three simple algorithms, derived by synchronous, probabilistic and asynchronous BN. The probabilistic and asynchronous algorithms have shown interesting behaviors.

The contribution of this work is twofold: first, we proved that it is possible to switch from the propositional space to the boolean network space, preserving the correctness of the results; second, BN-computation provides a general framework for local search procedures.

We are currently working on the introduction of heuristics into BNs algorithms; future works concern the design of more complex mappings and the extension of BNs model.

## Appendix A

**Proposition 1**: the mapping $\mu 1$ between a SAT instance $\Phi$ (with $n$ variables) and the corresponding BN $\Re$ of $n$ nodes is sound and complete.
*Proof*:
1- completeness) Suppose that $(x_1, \ldots, x_n)$ is a solution of $\Phi$, then all the clauses are satisfied, that is $c_1 = c_2 = \ldots = c_m = 1$. All the boolean function are: $F_i = (1 \wedge \ldots \wedge 1 \wedge x_i) \vee 0 \vee \ldots \vee 0 \Leftrightarrow F_i = x_i$, this is equivalent to a dynamics given by the evolution equations $x_i(t+1) = x_i(t)$ $\forall i$ $(1 \leq i \leq n)$ corresponding to a fixed point trajectory.
2- soundness) The current hypothesis is that $(x_1, \ldots, x_n)$ is a fixed point for $\Re$; suppose, *ab absurdo*, that $(x_1, \ldots, x_n)$ is not a satisfying assignment for $\Phi$; therefore a non satisfied clause $c_j$ exists, i.e., $c_j = 0$. Take a variable $x_i$ such that $x_i$ belongs to $c_j$; there are two cases:
a) $c_j = (\ldots \vee x_i \vee \ldots) = 0$, this implies that $x_i = x_i(t) = 0$ and then all the clauses containing literal $\sim x_i$ are satisfied; we have $F_i = x_i(t+1) = (1 \wedge x_i) \vee \ldots \vee \sim c_j = 0 \vee \sim c_j = 1$, that is $x_i(t) \neq x_i(t+1)$ and this contradicts the hypothesis.
b) $c_j = (\ldots \vee \sim x_i \vee \ldots) = 0$, this implies that $x_i = x_i(t) = 1$ and then all the clauses which contain the literal $x_i$ are satisfied; we have $F_i = x_i(t+1) = (\ldots \wedge c_j \wedge x_i) \vee 0 \vee \ldots \vee 0 = (0 \wedge 1) \vee 0 = 0$, that is $x_i(t) \neq x_i(t+1)$ and this contradicts the hypothesis.

## Appendix B

**Proposition 2**: If the PBN (ABN) is generated by means of $\mu 1$ from a satisfiable boolean formula, for every initial state the probability that the network reaches a fixed point tends to 1.
*Proof* (sketched): to prove the proposition we need to use some results about Markov chains (MCs) ([Feller, 1968]). The main result we use is the following: given a MC, if $C$ is the closed set given by all the persistent states of the MC, the chain will eventually reach $C$ with probability 1.

**Obs1**: The state space trajectory described by a PBN (ABN) is a MC.

**Lemma1**: If $\mathbf{x}^*$ is a fixed point, then $\mathbf{x}^*$ is an *absorbing state* for the MC (that is $\mathbf{x}^*$ is an irreducible set of only one state).
*Proof*: it is easy to prove that the sum of the probabilities of all the transitions from any node is 1 (by using combinatorial analysis). Since $\mathbf{x}^*$ is a fixed point, for i=1,...,*n* is $x_i(t+1) = F_i = x_i(t)$. Then, each variable has probability 1 to maintain the old value and the only transition is represented by a self-arc, which has probability 1. Each fixed point can communicate only with itself, then it is an absorbing state.

**Lemma2**: The states which are not fixed points are *transient states*, in the MC sense.
*Proof*: we will prove that a state $\mathbf{x}$, which is not a solution, communicates with a solution $\mathbf{s}$; then, since solutions communicate only with themselves, $\mathbf{x}$ is a transient state. We will prove that there exists a path (constituted by 1-Hamming transitions), with positive probability, between $\mathbf{x}$ and $\mathbf{s}$. The trajectory is the result of two overlapped mechanisms: the dynamics mechanism and the functional mechanism. The first quantifies the probability of any transition, while the second specifies which transitions are allowed and which are not. Suppose that the network has only one fixed point $\mathbf{s} = (s_1,...,s_n)$, $\mathbf{x} = (x_1,...,x_n)$, $\mathbf{x} \neq \mathbf{s}$ and define $V=\{x_1,...,x_n\}$, $I = \{x_i \in V \text{ s.t. } x_i \neq s_i\}$. Since $\mathbf{x}$ is not a solution, there exists a subset $\Gamma$ of unsatisfied clauses. There is, at least, one $x_i \in I$ such that $x_i \in c_j \in \Gamma$, then the transition $(x_1,..,x_i,..,x_n) \rightarrow (x_1,...,\sim x_i,..,x_n)$ is allowed (by the functional mechanism). If such transition does not exist, $\Gamma$ would be constituted only by clauses involving variables belonging to V\I; but those variables have a value that satisfies the formula, then also the subset $\Gamma$ and this is a contradiction. The probability of the transition (given by the dynamics mechanism) is $p(1-p)^{n-1}$ for the PBN and $1/n$ for the ABN. By iterating this step it is possible to reach $\mathbf{s}$ with a succession of transitions, each one obtained by a single variable update. The probability of this path is the product of the transition probabilities and it is positive. Suppose, now, that the network has a set S of fixed points, with |S|>1. If we take $\mathbf{s} \in S$ and $\mathbf{x} \notin S$ we can repeat the previous proof. Thus every $\mathbf{x} \notin S$ can reach every $\mathbf{s} \in S$ with positive probability.
*Conclusion*: the states of the network can be represented by $E = T \cup \{\mathbf{s_1}\} \cup ... \cup \{\mathbf{s_h}\}$, where T is the set of transient states, and $\{\mathbf{s_1}\} \cup ... \cup \{\mathbf{s_h}\} = C$ is the closed set of the absorbing states. A theorem states that the MC reaches *C* with probability 1.